\documentclass[letterpaper, 10 pt, conference]{ieeeconf}  

\IEEEoverridecommandlockouts                              

\usepackage[utf8]{inputenc}
\usepackage{graphicx}       
\usepackage{amsmath}        
\usepackage{amssymb}
\usepackage{bm}
\usepackage{booktabs}       
\usepackage{multirow}
\usepackage{subcaption}
\usepackage{algorithm}
\usepackage[noend]{algpseudocode}
\usepackage{cite}
\usepackage{url}
\usepackage{xcolor}
\usepackage[colorlinks=true, allcolors=black, hidelinks]{hyperref}
\usepackage{cleveref}       
\usepackage{placeins}       
\usepackage{stfloats}       

\graphicspath{{pics/}{figures/}}

\newcommand{\dsname}{LAPSE}

\title{\LARGE \bf
LAPSE: An Egocentric Indoor Benchmark for Continual Visual Place Recognition across Scenes and Illumination
}

\author{Guangyu Li$^{1,*}$, Peijun Dong$^{2,*}$, 
Xiangze Meng$^{3}$, Xiao Li$^{4}$, Yina Jian$^{5}$, 
Tao Tang$^{1}$, Chio-In Ieong$^{1}$, 
and Mingkun Xu$^{1,\dagger}$%
\thanks{$^{1}$ Guangdong Institute of Intelligence Science and Technology, China.}%
\thanks{$^{2}$ Cornell University, USA.}%
\thanks{$^{3}$ Jilin University, China.}%
\thanks{$^{4}$ State Key Laboratory of Deep-Sea Science and Intelligent Technology, 
Institute of Deep-Sea Science and Engineering, Chinese Academy of Sciences, China.}%
\thanks{$^{5}$ Columbia University, USA.}%
\thanks{$^{*}$ Guangyu Li and Peijun Dong contributed equally to this work.}%
\thanks{$^{\dagger}$ Corresponding author: Mingkun Xu 
(e-mail: xumingkun@gdiist.cn).}%
}

\begin{document}

\maketitle
\thispagestyle{empty}
\pagestyle{empty}

\begin{abstract}
Visual place recognition (VPR) is a core problem in robot navigation, yet
indoor VPR remains under-studied even as mobile robots proliferate indoors.
Reliable metric ground truth is a major practical obstacle: GNSS is generally
unavailable indoors, while motion capture is typically limited to
instrumented areas. Existing indoor VPR benchmarks include long-term revisits
under varying lighting, but offer limited support for studying paired sessions
under light and dark conditions separately from scene changes. We present
\dsname{}, an indoor VPR benchmark collected autonomously by a quadruped with
an egocentric camera across five everyday scenes, four with separate
traversals under light and dark conditions, plus a corridor recorded only
under the light condition. Ground-truth poses are obtained by relocalizing
every traverse to a prior LiDAR map of its scene; trajectory accuracy is
assessed against an independent AprilTag reference in one room-scale area.
Across thirteen methods spanning 2016--2026, results are tightly clustered
for same-scene, same-condition retrieval, whereas light-to-dark cross-session
retrieval has lower accuracy and more than twice the between-method spread.
The scene and lighting-condition axes further support a continual VPR
benchmark in which environments and lighting conditions arrive in sequence.
The dataset and evaluation code will be released.
\end{abstract}

\section{INTRODUCTION}

Visual place recognition (VPR) lets a mobile robot recover where it is from
an image alone and is widely used as a front end for relocalization in
navigation~\cite{lowry_survey,vpr_tutorial}. Much of large-scale VPR
benchmarking has focused on outdoor environments, partly because GPS/INS
makes metric ground truth easier to obtain. Indoors, GNSS is generally
unavailable, while motion capture is typically limited to instrumented
areas. Collecting repeated, metrically aligned indoor traversals at building
scale therefore requires dedicated mapping and localization infrastructure.

This gap matters increasingly as quadrupeds, humanoids, and wheeled service
robots move into indoor spaces. Metre-scale rooms demand tighter retrieval
tolerances than outdoor routes. A quadruped's low camera height limits scene
context, while its gait adds viewpoint variation and motion blur. Nearby
pedestrians can dominate the image, and indoor lighting can change abruptly.

\begin{figure}[!t]
    \centering
    \includegraphics[width=\columnwidth]{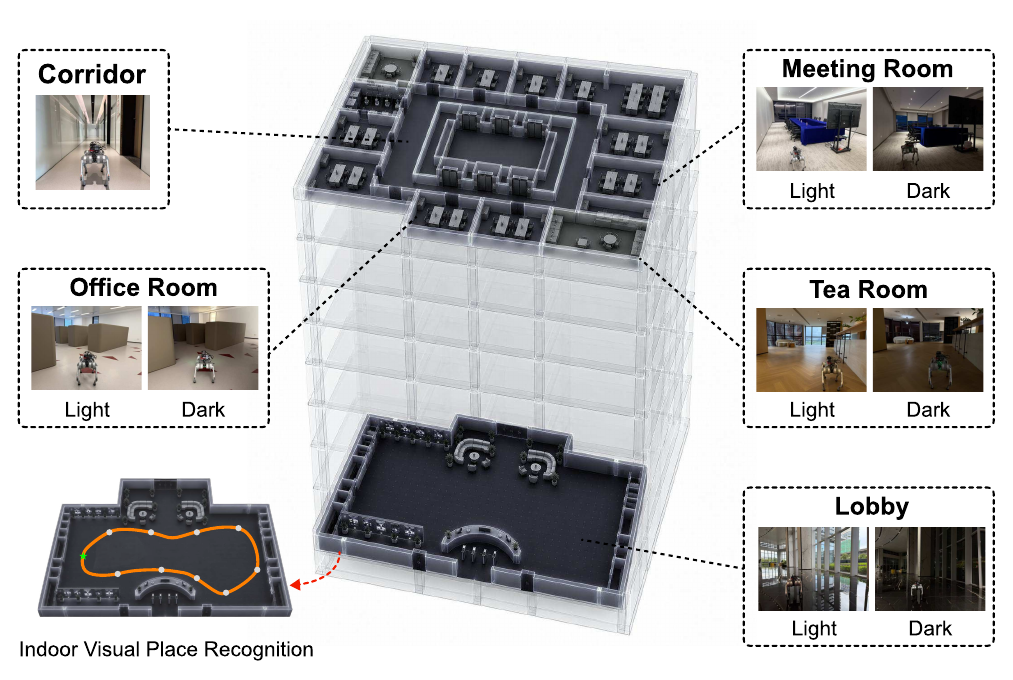}
    \caption{\dsname{}: five indoor scenes recorded autonomously by a
    quadruped, four under both light and dark conditions and the corridor
    under the light condition only.}
    \label{fig:teaser}
\end{figure}

We present \dsname{}, an indoor VPR benchmark recorded autonomously by a
quadruped carrying an egocentric camera and a LiDAR--inertial odometry stack
(\cref{fig:teaser}). Four everyday scenes are traversed under both light and
dark conditions, while the corridor is recorded only under the light
condition. This yields nine scene--condition pairs, hereafter called
\emph{cells}. Metric poses require no installed positioning infrastructure:
every traverse is relocalized into a prior LiDAR map of its scene. Trajectory
accuracy is assessed against an independent AprilTag reference in one
room-scale area. Because the cells are organized along a scene axis and a
lighting-condition axis, the same recordings support a static retrieval
benchmark, a light-to-dark cross-session benchmark, and a continual VPR
benchmark.

\noindent Our contributions are:

\begin{itemize}
  \item \textbf{An autonomously collected indoor VPR dataset and alignment
        protocol.} \dsname{} contains five everyday indoor scenes recorded
        by a quadruped, including paired sessions under light and dark
        conditions in four scenes. Each scene is mapped once, and its
        repeated traversals are relocalized into the same LiDAR-map frame.
        We assess trajectory accuracy against an independent AprilTag
        reference in one room-scale area and quantify cross-traverse
        repeatability.

  \item \textbf{A static benchmark of same-condition and cross-condition
        indoor retrieval.} Across thirteen methods spanning 2016--2026,
        the Recall@1 spread is $9.6$ points for same-scene, same-condition
        retrieval and $21.0$ points for light-to-dark cross-session
        retrieval, with different method orderings across the two protocols.

  \item \textbf{A dual-incremental benchmark design for continual indoor
        VPR.} Sequencing the nine scene--condition pairs along scene and
        lighting-condition axes supports separate evaluation of adaptation
        to new environments and revisits under changed lighting.
\end{itemize}

\section{RELATED WORK}
\label{sec:related}
%

Visual place recognition retrieves geo-referenced images to provide place
hypotheses for loop closure and relocalization~\cite{lowry_survey,vpr_tutorial}.
A false loop-closure match can propagate into the map, so global descriptors
must preserve place identity despite changes in viewpoint, season and
illumination. Continual VPR extends this requirement across deployment: the
model must adapt to new environments without forgetting places learned
earlier.

\textbf{Ground truth indoors.}
Outdoors, GPS/INS can place repeated traversals in a common coordinate frame,
although signal loss and drift can compromise pose accuracy~\cite{robotcar}.
Indoors, GNSS is generally unavailable, while motion capture requires an
instrumented area. Indoor localization datasets therefore use mapping
pipelines: Baidu Mall registers images to a LiDAR point cloud~\cite{baidumall};
NAVER LABS obtains initial poses from LiDAR SLAM and refines them with
structure from motion~\cite{naverindoor}; and InLoc provides manually verified
query poses in the coordinate frame of a 3D map~\cite{inloc}. We separately
assess our SLAM trajectory against an AprilTag reference using absolute
trajectory error in one validation room~\cite{apriltag,ate_tum}.

\textbf{Visual place recognition.}
NetVLAD~\cite{netvlad} and GeM~\cite{gem_pool} are established global
aggregation methods. CosPlace~\cite{cosplace} and
EigenPlaces~\cite{eigenplaces} train with geographically defined classes,
while GSV-Cities~\cite{gsvcities} provides labeled Street View data used
to train MixVPR~\cite{mixvpr} and BoQ~\cite{boq}.
AnyLoc~\cite{anyloc} aggregates frozen DINOv2~\cite{dinov2} features
without VPR-specific training. SALAD~\cite{salad}, MegaLoc~\cite{megaloc},
FoL~\cite{fol}, ImAge~\cite{image} and SAGE~\cite{sage} also build on
DINOv2, using different descriptor-construction or adaptation strategies.
Local features can further re-rank global retrievals, as in
Patch-NetVLAD~\cite{patchnetvlad} and EffoVPR~\cite{effovpr}.
Recent indoor work covers year-long appearance variation in crowded
scenes~\cite{nycindoorvpr}, text-assisted retrieval in repetitive multi-floor
interiors~\cite{textinplace}, and multimodal indoor/outdoor place recognition
~\cite{yu2023brain}. \dsname{} complements these settings by pairing light
and dark sessions within scenes and organizing scene and lighting-condition
changes as separate continual-VPR axes.

\textbf{Continual VPR.}
Continual learning uses rehearsal~\cite{er}, parameter
regularization~\cite{ewc,si}, and distillation~\cite{lwf}. AirLoop~\cite{airloop}
combines a within-environment triplet buffer with relational memory-aware
synapses (RMAS) and relational knowledge distillation (RKD), both aimed at
preserving pairwise descriptor structure. InCloud~\cite{incloud} uses
angular structure-aware distillation and a relaxation schedule for LiDAR
place recognition. VIPeR~\cite{viper} adaptively mines positive and negative
samples and combines sensory, working and long-term memory with RMAS and
probabilistic knowledge distillation for visual place recognition.
AirLoop and VIPeR evaluate scene increments on simulated TartanAir and
appearance increments on RobotCar (lighting) and Nordland (seasons)
~\cite{tartanair,robotcar,nordland}. \dsname{} evaluates both increment types
within one real indoor benchmark; task-level transfer-matrix
summaries~\cite{gem} alone do not attribute performance changes to either
axis.

\section{METHOD}
\label{sec:method}

\begin{figure*}[t]
    \centering
    \includegraphics[width=0.82\textwidth]{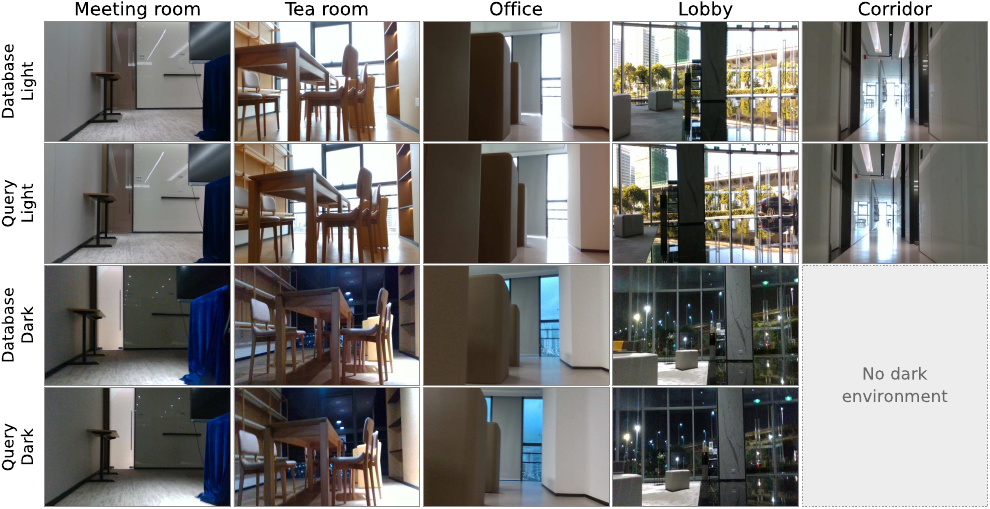}
    \caption{The nine cells of \dsname{}. Columns are the five scenes; rows
    show the database and query traverses under the light condition (top two)
    and the dark condition (bottom two). The corridor is available only under
    the light condition.}
    \label{fig:dataset}
\end{figure*}

\subsection{Collection Protocol}
\label{sec:protocol}








\textbf{1) Platform and sensor suite.}
We use a Unitree Go2 quadruped equipped with an NVIDIA Jetson Orin NX 16G
and a head-mounted Intel RealSense D435i RGB-D camera. The computing module
is fixed to the robot using a custom 3D-printed mount. Camera extrinsics are 
obtained through an AprilTag-based hand--eye
calibration, providing the rigid transform between the camera and robot body frames.

\textbf{2) Quadruped motion characteristics.}
Quadruped locomotion introduces gait-induced vertical motion, roll, and pitch,
causing additional viewpoint variation and motion blur compared with more
stable wheeled platforms. Even with the same planned route, repeated traversals
cannot exactly reproduce the same physical trajectory because of gait and
path-tracking variations. These deviations preserve overall spatial
correspondence while introducing realistic viewpoint changes between
database/reference and query traversals. To limit motion blur, we keep the robot's motion slow and steady. Its relatively low camera height also
produces a low egocentric viewpoint with reduced scene context.

\textbf{3) Data Acquisition procedure.}
For a new scene, we first build a 3D point-cloud map with the Go2's built-in
mapping system. Because LiDAR mapping is insensitive to illumination, this one
map serves every later traverse of that scene, under either condition.

An agent on the edge computing device reduces the point-cloud map to a 2D occupancy map by rasterisation,
height filtering and obstacle inflation, plans a coverage path over it, and
drives the robot along that route through ROS~2 while recording the sensor and
robot-state streams.

Repeated collection in the same scene reuses that map and route, with the
robot redeployed to the same or an adjacent starting area. Coverage therefore
stays consistent across traverses while the gait-induced deviations do not
repeat, so repeated traversals of a scene form an independent database/query pair
rather than frame-level duplicates. A cross-illumination traverse follows the
same procedure---same map, same route, matched starting area---under a different lighting condition.

\subsection{Ground-Truth Construction}
\label{sec:gt}

\textbf{Pose source.}
Within each scene, every traverse is relocalized into one prior
LiDAR-inertial map of that scene, built offline by an independent mapping
run. Poses therefore share a frame by construction, and we apply no trajectory registration.
 
\textbf{Timestamps and association.}
Every message carries a reception timestamp and a ROS message-header
timestamp (\texttt{header.stamp}). We use the header timestamp for color
frames and the reception timestamp for odometry, as these better approximate
acquisition time in their respective streams. A color frame is kept if a
depth frame falls within $40$\,ms and two odometry samples bracket it within
$200$\,ms; its pose is then interpolated linearly between them, with yaw
interpolated circularly.
 
\textbf{Labels.}
A database image is a positive for a query if it lies in the same scene with
$d \le 0.5$\,m and $|\Delta\psi| \le 40^\circ$; $0.5 < d \le 1.0$\,m is
\emph{ignored}. The same-scene constraint plays the role of the scene-ID
constraint of ScanNetPR~\cite{scannetpr}: a geometrically close image from another
scene is never a positive. Queries the database does not cover have no
positive and are removed; removal rates range from $0.00$ to
$11.25\,\%$ across cells.
 
\textbf{Thresholds.} \Cref{tab:threshold} summarizes the coverage,
scene-scale, repeatability, and navigation considerations behind the
$0.5$\,m positive radius. We also require $|\Delta\psi|\le40^\circ$,
following MSLS~\cite{msls}, to exclude nearby views facing opposite
directions along adjacent coverage-path scan lines.

\raggedbottom
\begin{table}[t]
\caption{Evidence informing the $0.5$\,m positive radius on \dsname{}.
Database-coverage counts use distance only; evaluation labels also impose
a $40^\circ$ yaw gate. Cross-traverse offsets measure repeatability, not
absolute pose accuracy.}
\label{tab:threshold}
\centering
\footnotesize
\setlength{\tabcolsep}{3pt}
\begin{tabular}{@{}p{1.9cm}p{1.35cm}p{3.45cm}@{}}
\toprule
Consideration & Indicated scale & Evidence \\
\midrule
Database coverage
  & $>0.25$\,m
  & At $0.25$\,m, 16\,\% of queries have no positive;
    at $0.5$\,m, 0\,\% do (distance only). \\
Scene extent
  & below $1.0$\,m
  & At $1.0$\,m, 86\,\% of queries already have
    $>10$ positives. \\
\midrule
Cross-traverse repeatability
  & offset $<0.09$\,m
  & Corresponding-location offsets span
    $0.006$--$0.088$\,m (\cref{sec:gtvalidation}). \\
Task semantics
  & $\approx 0.5$\,m
  & Same place at navigation scale: $0.7$\,m body length
    and $0.25$--$0.35$\,m waypoint tolerance. \\
\bottomrule
\end{tabular}
\end{table}

\subsection{Dual-Incremental Dataset Design: Scene $\times$ Illumination}
\label{sec:design}

\dsname{} targets long-term indoor relocalization, where a robot revisits
the same building and both the places it knows and their appearance change
between visits. We therefore structure it along two incremental axes,
\emph{scene} and \emph{lighting condition}. Each cell is a scene--condition pair with its own database set and a disjoint query traverse.
Four scenes provide both light and dark conditions, while the corridor is
recorded only under the light condition, giving \textbf{nine} cells
(\cref{fig:dataset}). The cell rather than the scene is the evaluation unit
because the same room under light and dark conditions defines two distinct
retrieval tasks; traverse identity is excluded from the task definition and
retained only in the metadata.
 
\textbf{Scene axis.}
Meeting room, tea room, office, corridor and lobby span a factor of $28$ in
floor area ($51$ to $1{,}419$\,m$^2$), so the axis varies scene scale as well
as scene identity.

\textbf{Illumination axis.}
The two conditions reflect normal building use: bright daytime traversals
and lower-light traversals beginning at dusk, rather than a scripted
lights-off experiment. Some indoor lighting remains in the latter, so
``dark'' denotes an evening, lower-light session, not complete darkness.
The appearance contrast varies by scene: the office remains similarly
bright with its lights on, while the meeting room becomes much dimmer.
The corridor shows little contrast between its daytime and evening
recordings, so it is represented only by the light condition and excluded
from cross-illumination evaluation.
 
\textbf{Realism and privacy.}
Traverses were recorded during normal building operation, retaining
passers-by, vehicles visible through the lobby windows, and changes in
daylight. These factors introduce natural appearance variation, especially
in public areas. For privacy, we locate faces using body keypoints and blur
the facial regions without discarding the images. Apart from face blurring,
we apply no other image-level edits or photometric adjustments to the
released RGB frames.

\begin{figure}[!htbp]
    \centering

    \begin{subfigure}[t]{0.48\columnwidth}
        \centering
        \includegraphics[width=\linewidth]{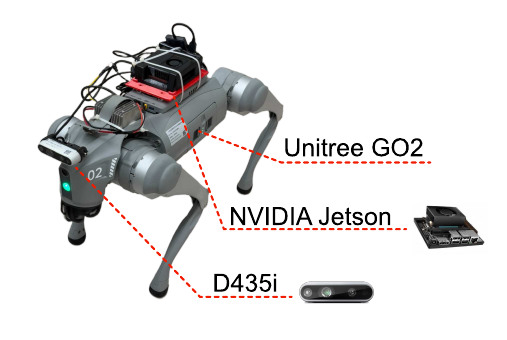}
        \caption{Robotic platform}
        \label{fig:gt_platform}
    \end{subfigure}
    \hfill
    \begin{subfigure}[t]{0.48\columnwidth}
        \centering
        \includegraphics[width=\linewidth]{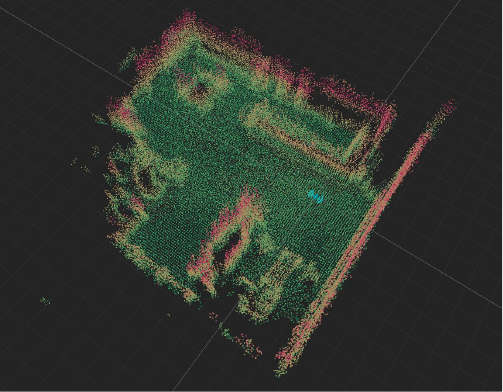}
        \caption{3D map}
        \label{fig:gt_pointcloud}
    \end{subfigure}

    \vspace{1.2mm}

    \begin{subfigure}[t]{0.48\columnwidth}
        \centering
        \includegraphics[width=\linewidth]{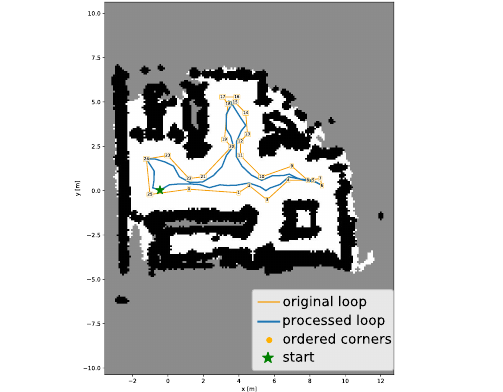}
        \caption{2D trajectory}
        \label{fig:gt_trajectory}
    \end{subfigure}
    \hfill
    \begin{subfigure}[t]{0.48\columnwidth}
        \centering
        \includegraphics[width=\linewidth]{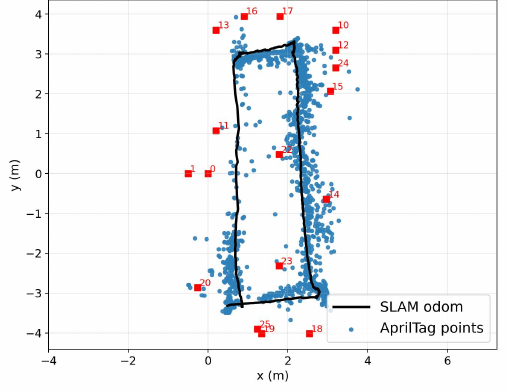}
        \caption{AprilTag validation}
        \label{fig:gt_apriltag}
    \end{subfigure}

    \caption{Ground-truth validation pipeline. 
    (a) Robotic platform.
    (b) 3D point-cloud reconstruction.
    (c) 2D map with the robot trajectory.
    (d) AprilTag-based validation of the estimated trajectory.}
    \label{fig:gt_validation}
\end{figure}

\section{EXPERIMENTS}
\label{sec:exp}

\subsection{Ground-Truth Validation}
\label{sec:gtvalidation}

The trajectory estimated by the onboard LiDAR-based SLAM system was validated
against an independent AprilTag reference (\cref{fig:gt_validation}). AprilTag
world poses were measured manually and were independent of the onboard
localization system. For each detected tag, the camera pose was recovered by
PnP and transformed to the robot body frame using the calibrated
camera-to-body extrinsics.

\textbf{Trajectory evaluation.}
The SLAM and AprilTag trajectories were aligned by a rigid-body transform,
scale fixed to one~\cite{umeyama}, and compared by Absolute Trajectory Error
(ATE)~\cite{ate_tum}: scale is observable for a range-inertial system, and a
free 7-DoF fit would absorb any residual into the alignment and understate the
ATE. We report the separate $\mathrm{Sim}(3)$ scale as a diagnostic:
$0.96\pm0.02$ in the validation room suggests a scale discrepancy of
order $4\,\%$. A rigid transform shared by database and query poses
cancels in pairwise distances, but scale does not; $4\,\%$ of the
$0.5$\,m threshold is $0.02$\,m.
Validation used nine loop traversals in an $8\,\mathrm{m}\times3\,\mathrm{m}$
pantry room with 17 AprilTags.

\begin{table}[!htbp]
\centering
\caption{Pantry-room GT accuracy measured by ATE against an independent
AprilTag reference, after rigid alignment with the scale fixed to one. The
$\mathrm{Sim}(3)$ row is the scale of a separate 7-DoF fit, reported as a
diagnostic and not applied. Values are mean $\pm$ std over the nine runs.}
\label{tab:pantry_2d_3d_ate_summary}
\begin{tabular}{lcc}
\toprule
 & 2D & 3D \\
\midrule
RMSE (m)        & $0.29 \pm 0.02$ & $0.34 \pm 0.03$ \\
Mean (m)        & $0.23 \pm 0.02$ & $0.28 \pm 0.02$ \\
Median (m)      & $0.20 \pm 0.03$ & $0.22 \pm 0.05$ \\
P90 (m)         & $0.41 \pm 0.05$ & $0.48 \pm 0.07$ \\
Max (m)         & $1.70 \pm 0.48$ & $1.90 \pm 0.43$ \\
\midrule
Sim(3) scale    & \multicolumn{2}{c}{$0.96 \pm 0.02$} \\
\bottomrule
\end{tabular}
\end{table}

\textbf{Interpretation.}
The lower 2D ATE indicates that vertical motion and attitude perturbations
contribute to the 3D trajectory error. Since VPR labels mainly depend on planar
place consistency, 2D error is more relevant to retrieval. Across repeated
traversals, corresponding locations differed by only $0.006$--$0.088\,\mathrm{m}$,
well below the $0.5\,\mathrm{m}$ positive radius. These offsets describe
cross-traverse repeatability, not absolute accuracy or a worst-case error
bound on individual labels. The AprilTag comparison measures absolute
pose error in the validation room; it does not directly measure the error
of every query--database separation.

\subsection{Visual Place Recognition}
\label{sec:vpr}

\subsubsection{Evaluation Protocols}

Retrieval is evaluated per cell (\cref{sec:design}). Within a cell the
database and query sets come from disjoint traverses, rather than frames split from one traverse;
temporally adjacent frames therefore cannot leak between them. Two
protocols are defined on this basis:

\begin{enumerate}
  \item \textbf{A --- same-condition.} Database and query are drawn from
      disjoint traversals of the same scene under the same nominal lighting
      condition. Nine tasks; this measures retrieval across repeated
      traversals without an intended change in lighting condition.
  \item \textbf{B --- cross-illumination.} The database is the light traverse
        and the query the dark traverse of the same scene. Four tasks, the
        corridor having no dark condition; this measures recognition across
        light-to-dark sessions.
\end{enumerate}

The two protocols share one set of labels and one evaluation script. Within
each scene, the traverses follow nominally matched routes under the specified
lighting conditions. Because they were recorded at different times, viewpoint,
path deviations, and transient scene content may also differ. We therefore
interpret the A--B gap as light-to-dark cross-session robustness, not the
isolated causal effect of illumination. 
The protocols cover different scene sets: protocol~A includes the corridor,
which has only light-condition data and therefore no protocol-B counterpart.

Each method is run at the input resolution its authors specify
($320$--$512$\,px); we do not re-tune it.
We report Recall@1/5/10/25, macro-averaged over cells: database size varies
$5.7\times$ across cells (84--482), so a query-weighted mean would let the
largest scenes dominate. Because the easiest cell is already
saturated---the best method reaches $100.0$ on
\texttt{meetingroom/light}---per-cell results are reported alongside the mean
rather than in place of it. Database images falling inside the ignore band
are skipped without consuming a rank; the
convention is not negligible, as charging them a rank lowers mean Recall@1 by
$7.18$ points.

The primary protocols use task-specific galleries: protocol~A searches
the database associated with the query's scene and lighting condition,
while protocol~B searches the light-condition database of the query's scene. They therefore measure within-scene
retrieval with scene identity given; blind building-wide localization
and retrieval with a growing gallery are outside our scope.

\subsubsection{Results}

\Cref{tab:vpr_main} reports Recall@$N$ for thirteen methods on both
protocols. On protocol~A, NetVLAD (2016) has the highest macro-averaged
Recall@1 at $90.9\,\%$, and the spread across methods is $9.6$ points.
All thirteen methods have lower reported Recall@1 on protocol~B, where
BoQ leads at $75.2\,\%$ and the spread reaches $21.0$ points. These
aggregate differences do not isolate the effect of illumination:
protocol~A covers nine cells, whereas protocol~B covers four scenes.

Method ordering also changes. NetVLAD has the highest Recall@1 on
protocol~A but ranks twelfth of thirteen on B at $56.3\,\%$. \Cref{tab:percell} further shows variation
among cells: under A, the mean Recall@1 across methods ranges from
$75.7\,\%$ in \texttt{corr/L} to $97.2\,\%$ in \texttt{meet/L}; under B,
it ranges from $54.4\,\%$ in the lobby to $83.4\,\%$ in the office.
The meeting room has the largest between-method spread on B
($39.1$ points).

\begin{table*}[t]
\caption{VPR baselines on \dsname{}: Recall@$N$ (\%), macro-averaged
over cells. ``Dim'' denotes descriptor dimension. Best and worst entries
are \textbf{bold} and \underline{underlined}; spreads use unrounded values.}
\label{tab:vpr_main}
\centering
\small
\setlength{\tabcolsep}{6pt}
\begin{tabular}{llccccccccc}
\toprule
\multirow{2}{*}{Method} & \multirow{2}{*}{Year} & \multirow{2}{*}{Dim}
 & \multicolumn{4}{c}{A -- same-condition}
 & \multicolumn{4}{c}{B -- cross-illumination} \\
\cmidrule(lr){4-7}\cmidrule(lr){8-11}
 & & & R@1 & R@5 & R@10 & R@25 & R@1 & R@5 & R@10 & R@25 \\
\midrule
NetVLAD~\cite{netvlad} & 2016 & 4096  & \textbf{90.9} & 96.9 & 98.2 & 98.8 & 56.3 & 83.5 & 89.8 & 94.9 \\
CosPlace-VGG16~\cite{cosplace} & 2022 & 512   & 88.1 & 95.9 & 97.8 & 99.0 & 63.2 & 84.1 & 89.9 & 96.3 \\
CosPlace-R50~\cite{cosplace} & 2022 & 2048  & 87.8 & 95.1 & 97.0 & 98.7 & 59.7 & 82.2 & 89.5 & 95.0 \\
MixVPR~\cite{mixvpr} & 2023 & 4096  & 90.1 & \textbf{96.9} & \textbf{98.3} & 99.6 & 72.7 & 89.3 & 95.5 & 97.6 \\
EigenPlaces~\cite{eigenplaces} & 2023 & 2048  & 89.4 & 96.8 & 98.3 & 99.4 & 64.8 & 88.2 & 91.5 & 97.1 \\
AnyLoc~\cite{anyloc} & 2023 & 49152 & 87.3 & 95.0 & 97.9 & 98.9 & 74.0 & 91.3 & 96.1 & 98.8 \\
SALAD~\cite{salad} & 2024 & 8448  & 87.9 & 95.4 & 97.0 & 99.1 & 71.3 & 90.2 & 96.6 & 98.9 \\
BoQ~\cite{boq} & 2024 & 12288 & 88.5 & 95.8 & 98.1 & \textbf{99.6} & \textbf{75.2} & \textbf{91.9} & \textbf{97.8} & \textbf{99.3} \\
MegaLoc~\cite{megaloc} & 2025 & 8448  & 89.0 & 96.4 & 98.3 & 99.5 & 67.0 & 87.6 & 92.4 & 96.9 \\
FoL-global~\cite{fol} & 2025 & 8448  & 85.6 & 95.2 & 97.1 & 98.6 & 69.2 & 86.3 & 91.9 & 96.8 \\
ImAge~\cite{image} & 2025 & 6144  & 87.7 & 95.8 & 97.6 & 99.1 & 69.6 & 85.6 & 90.0 & 95.5 \\
SAGE~\cite{sage} & 2026 & 8448  & 87.2 & 95.2 & 97.1 & 98.3 & 63.9 & 86.8 & 92.1 & 96.1 \\
DINOv2 (raw)~\cite{dinov2} & --  & 768   & \underline{81.3} & \underline{92.9} & \underline{95.3} & \underline{97.6} & \underline{54.2} & \underline{77.5} & \underline{84.5} & \underline{90.7} \\
\midrule
\multicolumn{3}{l}{\textbf{Spread}} & \textbf{9.6} & 4.0 & 3.1 & 2.1 & \textbf{21.0} & 14.4 & 13.3 & 8.6 \\
\bottomrule
\end{tabular}
\end{table*}

\begin{table}[!t]
\caption{Per-cell difficulty. Mean and spread over the thirteen methods of
\cref{tab:vpr_main}; $N_{\mathrm{db}}$ is the database size.}
\label{tab:percell}
\centering
\footnotesize
\setlength{\tabcolsep}{3.5pt}
\begin{tabular}{@{}lc cc cc cc@{}}
\toprule
\multirow{2}{*}{Cell} & \multirow{2}{*}{$N_{\mathrm{db}}$}
 & \multicolumn{2}{c}{R@1} & \multicolumn{2}{c}{R@5}
 & \multicolumn{2}{c}{R@10} \\
\cmidrule(lr){3-4}\cmidrule(lr){5-6}\cmidrule(lr){7-8}
 & & mean & spr. & mean & spr. & mean & spr. \\
\midrule
\multicolumn{8}{@{}l}{\emph{A --- same-condition}} \\
meet/L   &   90 &  97.2 &   9.0 &  99.7 &   1.5 &  100.0 &  0.0 \\
meet/D   &   84 &  89.1 &   9.9 &  93.0 &   5.6 &  94.5 &   4.2 \\
tea/L    &   96 &  95.8 &   7.8 &  99.7 &   1.1 &  99.9 &   1.1 \\
tea/D    &   97 &  89.4 &   7.8 &  97.5 &   3.3 &  99.1 &   4.4 \\
lobby/L  &  482 &  81.8 &  19.3 &  93.6 &  10.2 &  96.1 &   6.9 \\
lobby/D  &  348 &  78.5 &  19.4 &  91.8 &  11.5 &  95.5 &  10.1 \\
off/L    &  101 &  89.6 &   8.6 &  96.2 &   5.4 &  98.6 &   3.2 \\
off/D    &   84 &  92.8 &   8.6 &  98.8 &   3.7 &  99.3 &   2.5 \\
corr/L   &  305 & \underline{75.7} &   4.9 &  90.5 &   6.1 &  94.9 &   4.9 \\
\midrule
\multicolumn{8}{@{}l}{\emph{B --- cross-illumination}} \\
meet     &   90 &  58.4 & \textbf{39.1} &  85.8 &  20.7 &  92.0 &  16.1 \\
tea      &   96 &  68.7 &  23.4 &  86.1 &  22.3 &  92.0 & \textbf{23.4} \\
lobby    &  482 & \underline{54.4} &  25.6 & \underline{79.1} & \textbf{23.7} & \underline{86.8} &  17.8 \\
off      &  101 & \textbf{83.4} &  23.8 & \textbf{95.1} &   6.2 & \textbf{97.7} &   5.0 \\
\bottomrule
\end{tabular}
\end{table}

\subsubsection{Selected Model Contrasts}

Across the selected checkpoint pairs in \cref{tab:contrasts}, the
absolute Recall@1 differences are larger on protocol~B than on A.
The CosPlace-R50 checkpoint trained on GSV-Cities~\cite{gsvcities}
exceeds its SF-XL~\cite{cosplace} counterpart by $13.03$ points on B,
although input resolution also changes from $512$ to $320$\,px.
The CosPlace VGG16-to-ResNet-50 comparison changes B by $-3.44$ points
and also changes descriptor dimension from $512$ to $2048$.
The GSV-Cities-to-multi-set SALAD comparison changes B by $-4.33$
points. These checkpoint comparisons show sensitivity to model and
training choices; they do not isolate the effects of training corpus,
data volume, input resolution, or architecture.

AnyLoc exceeds raw DINOv2 by $19.8$ Recall@1 points on protocol~B.
This compares their complete feature-extraction and aggregation
pipelines; the difference cannot be attributed to aggregation alone.

\begin{table}[t]
\caption{Selected model contrasts on \dsname{}. $\Delta$A and
$\Delta$B are changes in macro-averaged Recall@1 (percentage points).
The SF-XL$\to$GSV-Cities pair also changes input resolution
($512\to320$\,px); the VGG16$\to$R50 pair also changes descriptor
dimension ($512\to2048$).}
\label{tab:contrasts}
\centering
\footnotesize
\setlength{\tabcolsep}{5pt}
\begin{tabular}{lcc}
\toprule
Contrast & $\Delta$A & $\Delta$B \\
\midrule
CosPlace: VGG16 $\to$ R50
    & $-0.28$ & $\mathbf{-3.44}$ \\
\midrule
CosPlace-R50: SF-XL $\to$ GSV-Cities
    & $+2.32$ & $\mathbf{+13.03}$ \\
SALAD: GSV-Cities $\to$ multi-set
    & $+1.07$ & $\mathbf{-4.33}$ \\
\bottomrule
\end{tabular}
\end{table}

\subsection{Continual Visual Place Recognition}
\label{sec:cl}

\subsubsection{Task Sequences}

\dsname{} is organized along two incremental axes---a new scene
(\emph{environment} increment) and a new illumination condition on a scene
already mapped (\emph{domain} increment)---from which we build three sequences
(\cref{tab:sequences}).

\textbf{Seq-ED} ($K = 9$) is primary: it traverses all nine cells and
exercises both axes at once. Its order follows the batches in which the robot
was commissioned---two scenes are taken on and covered under both conditions,
then the next two are added---so the interleaving of the two axes was fixed by
the collection schedule rather than chosen after seeing results. Within a
batch the order is the recording order up to one adjacent swap.

\textbf{Seq-E} ($K = 5$) holds illumination fixed and adds only scenes.
\textbf{Seq-D} ($K = 8$) first maps four scenes under the light condition
and then revisits the same four under the dark condition; its first half is
a scene increment, while only its second half exercises the
lighting-condition axis.
Seq-E and Seq-D are conceptual analogues of the scene-incremental
TartanAir and appearance-incremental RobotCar settings used by
VIPeR~\cite{viper,tartanair,robotcar}; their scores are not directly
comparable across datasets and label definitions.

\begin{table}[t]
\caption{The continual-learning sequences; L and D denote a scene's light
and dark conditions. The corridor, recorded only under the light condition,
appears only in Seq-ED and Seq-E.}
\label{tab:sequences}
\centering
\footnotesize
\setlength{\tabcolsep}{4pt}
\begin{tabular}{@{}llp{5.4cm}@{}}
\toprule
Sequence & $K$ & Order \\
\midrule
Seq-ED & 9 & tea/D, meet/D, meet/L, tea/L, off/D, lobby/D, off/L, corr/L,
             lobby/L \\
\addlinespace[2pt]
Seq-E  & 5 & meet/L, tea/L, off/L, corr/L, lobby/L \\
\addlinespace[2pt]
Seq-D  & 8 & meet/L, tea/L, off/L, lobby/L, meet/D, tea/D, off/D, lobby/D \\
\bottomrule
\end{tabular}
\end{table}

\textbf{Methods.} AirLoop~\cite{airloop} combines two components, both made
\emph{relational} so that they constrain pairwise embedding similarity rather
than individual outputs---the form a metric-learning task requires: relational
memory-aware synapses (RMAS), adapted from MAS, and relational
knowledge distillation (RKD).
Replay~\cite{er} keeps one buffer that persists across environments, so mined
triplets mix old and new; AirLoop's own buffer is rebuilt per environment.
InCloud$^{*}$~\cite{incloud} and VIPeR$^{*}$~\cite{viper} adapt the
respective structure-aware and probabilistic distillation objectives
to our common visual framework. Each replaces AirLoop's RKD; the
starred results do not represent the complete original systems.

\subsubsection{Metrics}

\textbf{Protocol.} Following AirLoop~\cite{airloop}, we train one environment
at a time in sequence order. For cell $j$, $D_j$ denotes its database set,
and $Q_j$ denotes a disjoint query
traverse. Training on cell $j$ uses $D_j$ and, where applicable, replayed
samples from earlier database sets; query images are not used for gradient
updates, triplet mining, or replay. At evaluation, $D_j$ also serves as the
retrieval gallery, and $R_{i,j}$ scores checkpoint $i$ by retrieving $Q_j$
against $D_j$, including cells not yet trained on. Thus training and gallery
images overlap by design, while queries come from disjoint traverses;
galleries remain cell-specific and fixed rather than pooled or growing.
The backbone is VGG-19 with GeM pooling~\cite{gem_pool}, 1024-D, trained
end-to-end.

\textbf{Evaluation.} AirLoop evaluates loop closure among images
within each environment~\cite{airloop}. Direct within-traverse
evaluation is poorly suited to our coverage paths: 12 of the 20
evaluation traverses have no qualifying loop pair. We therefore
use cross-traverse retrieval under protocol~A
(\cref{sec:vpr}) to define $R_{i,j}$.

\textbf{Scores.} For each query, we rank the database images,
skip ignored candidates, and count positives before the first
negative. Dividing by the query's total number of positives
gives recall at 100\,\% precision (R@100P); averaging over
$Q_j$ gives $R_{i,j}$. Ground-truth labels determine the
stopping point for evaluation, not a deployment threshold.
We summarize the $R_{i,j}$ matrix using AirLoop's AP, BWT,
and FWT averages:
\begin{equation}
  \mathrm{AP} = \frac{2}{K(K+1)}\sum_{i=1}^{K}\sum_{j=1}^{i} R_{i,j},
  \label{eq:ap}
\end{equation}
\begin{equation}
  \mathrm{BWT} = \frac{2}{K(K-1)}\sum_{i=2}^{K}\sum_{j=1}^{i-1}
                 \bigl(R_{i,j}-R_{j,j}\bigr),
  \label{eq:bwt}
\end{equation}
\begin{equation}
  \mathrm{FWT} = \frac{2}{K(K-1)}\sum_{i=1}^{K-1}\sum_{j=i+1}^{K} R_{i,j}.
  \label{eq:fwt}
\end{equation}

Negative BWT is forgetting. The triangular form weighs every step of
training, whereas the last-row convention reads only the final checkpoint.
All methods share one per-environment budget.

\begin{table*}[t]
\caption{Continual VPR on \dsname{}, recall at 100\,\% precision (AirLoop
convention). Seq-ED is primary. Entries with $\pm$ are mean\,$\pm$\,std over
three seeds; $^{*}$ denotes our adaptation of the cited distillation objective. 
Joint has no transfer by
construction. Best per column \textbf{bold} (Joint excluded), worst
\underline{underlined}.}
\label{tab:cl_main}
\centering
\small
\setlength{\tabcolsep}{3.5pt}
\begin{tabular}{lccccccccc}
\toprule
\multirow{2}{*}{Method}
 & \multicolumn{3}{c}{Seq-ED \emph{(primary)}} & \multicolumn{3}{c}{Seq-E}
 & \multicolumn{3}{c}{Seq-D} \\
\cmidrule(lr){2-4}\cmidrule(lr){5-7}\cmidrule(lr){8-10}
 & AP\,$\uparrow$ & BWT\,$\uparrow$ & FWT\,$\uparrow$
 & AP\,$\uparrow$ & BWT\,$\uparrow$ & FWT\,$\uparrow$
 & AP\,$\uparrow$ & BWT\,$\uparrow$ & FWT\,$\uparrow$ \\
\midrule
Finetune          & 64.25\,{\scriptsize$\pm$\,2.42} & $-6.67$\,{\scriptsize$\pm$\,2.65} & 57.78\,{\scriptsize$\pm$\,1.06} & 66.90 & $-5.82$ & 53.75 & 67.95 & $-3.12$ & 61.90 \\
Replay~\cite{er} & 65.31\,{\scriptsize$\pm$\,1.15} & $-5.86$\,{\scriptsize$\pm$\,0.52} & \underline{56.90}\,{\scriptsize$\pm$\,0.72} & \underline{66.79} & $\underline{-7.15}$ & 54.97 & 67.91 & $\underline{-6.14}$ & \underline{61.42} \\
EWC~\cite{ewc} & 65.64 & $-6.02$ & 58.85 & 67.25 & $-5.32$ & 53.92 & 66.91 & $-4.61$ & 61.73 \\
SI~\cite{si} & \underline{63.69} & $\underline{-7.53}$ & 58.09 & 67.71 & $-4.50$ & \underline{53.70} & \underline{66.86} & $-3.65$ & 61.70 \\
AirLoop~\cite{airloop} & 69.50\,{\scriptsize$\pm$\,1.32} & $-2.11$\,{\scriptsize$\pm$\,0.96} & 59.56\,{\scriptsize$\pm$\,0.68} & 68.78 & $-3.72$ & 55.16 & 70.05 & $-2.12$ & 62.47 \\
InCloud$^{*}$~\cite{incloud} & \textbf{69.78}\,{\scriptsize$\pm$\,0.61} & $\mathbf{-0.06}$\,{\scriptsize$\pm$\,0.32} & \textbf{62.40}\,{\scriptsize$\pm$\,0.55} & 70.70 & $-0.26$ & 55.86 & \textbf{70.68} & $-0.40$ & \textbf{63.63} \\
VIPeR$^{*}$~\cite{viper} & 69.12\,{\scriptsize$\pm$\,1.22} & $-0.27$\,{\scriptsize$\pm$\,0.34} & 62.28\,{\scriptsize$\pm$\,0.22} & \textbf{71.16} & $\mathbf{+0.39}$ & \textbf{56.42} & 70.35 & $\mathbf{-0.34}$ & 62.99 \\
\midrule
Joint             & 70.22 & -- & -- & 71.19 & -- & -- & 70.23 & -- & -- \\
\multicolumn{1}{l}{\textbf{Spread} (excl.\ Joint)} & \textbf{6.09} & & & \textbf{4.37} & & & \textbf{3.82} & & \\
\bottomrule
\end{tabular}
\end{table*}

\Cref{tab:cl_main} reports the continual setting. It uses R@100P, the
per-query recall at full precision of AirLoop and VIPeR, so its values are not
comparable with the R@1 of \cref{tab:vpr_main}; within the setting, a
$6.09$-point spread in average performance (AP, \cref{eq:ap}) on the
primary sequence shows the
benchmark is far from saturated. The two adapted variants obtain BWT of $-0.06$ (InCloud$^{*}$) and
$-0.27$ (VIPeR$^{*}$), versus $-6.67$ for naive fine-tuning.
Their BWT results also vary less across seeds---the standard deviation over three seeds is $0.32$ and
$0.34$ on BWT against $0.61$ and $1.22$ on AP. Seed-wise paired comparison
confirms the gain over AirLoop ($\Delta$BWT $+2.05 \pm 0.82$ and
$+1.84 \pm 0.65$, sign-consistent across all three seeds), whereas rehearsal
alone remains indistinguishable from naive fine-tuning ($\Delta$AP
$+1.06 \pm 1.28$, $\Delta$BWT $+0.82 \pm 2.13$, both reversing sign between
seeds). AirLoop, InCloud$^{*}$ and VIPeR$^{*}$ obtain Seq-ED AP of
$69.12$--$69.78$, close to the Joint reference of $70.22$. Because Joint is
reported without across-seed variability, this is a descriptive comparison.

\section{DISCUSSION}
\label{sec:discussion}

\textbf{Implications for indoor robots.}
On protocol~A, the tested descriptors are tightly clustered; protocol~B
separates them more clearly, and the two rankings show little association
among these methods. The selected checkpoint
contrasts suggest that training setup matters, although they do not
isolate individual causes. In continual evaluation, the adapted variants
approach the Joint AP and show less negative BWT than naive fine-tuning.
These results concern representation learning with fixed, task-specific
galleries; retrieval as a gallery grows across scenes remains untested.
Replay shows no consistent gain over fine-tuning across the three seeds.

\textbf{Limitations.}
\dsname{} covers one building and one robot platform, and protocol~B
includes only four scenes. Our scene and illumination sequences assess generalization to new environments
and retention under continual learning. Relocalization among many
visually similar rooms, such as offices with repeated layouts, and
floor identification in multi-storey buildings remain outside this
evaluation; topological or floor-level cues may help in these settings.
Compared with many outdoor VPR settings, indoor robots often move
closer to walls and other obstacles. At a tight corner, a nearby
surface may fill the camera view, leaving too little distinctive
visual information for reliable single-image retrieval. Such views
are uncommon in \dsname{}, but warrant focused evaluation in future
indoor VPR work.

\section*{ACKNOWLEDGMENT}
The building schematic in \cref{fig:teaser} was drawn by the authors from the
layout of the recording site and rendered in three dimensions with GPT 5.6;
it is schematic and not to scale. No other figure contains AI-generated content.



%

\FloatBarrier

\bibliographystyle{IEEEtran}
\bibliography{IEEEabrv,references}

@STRING{IEEE_J_PAMI       = "{IEEE} Trans. Pattern Anal. Mach. Intell."}

@STRING{IEEE_J_RAL        = "{IEEE} Robot. Autom. Lett."}

@STRING{IEEE_J_RO         = "{IEEE} Trans. Robot."}

@STRING{IEEE_M_RA         = "{IEEE} Robot. Autom. Mag."}

@inproceedings{netvlad,
  author    = {Arandjelovi\'{c}, Relja and Gronat, Petr and Torii, Akihiko and
               Pajdla, Tomas and Sivic, Josef},
  title     = {{NetVLAD}: {CNN} Architecture for Weakly Supervised Place Recognition},
  booktitle = {Proc. IEEE Conf. Comput. Vis. Pattern Recognit. (CVPR)},
  year      = {2016},
  pages     = {5297--5307},
  doi       = {10.1109/CVPR.2016.572}
}

@inproceedings{cosplace,
  author    = {Berton, Gabriele and Masone, Carlo and Caputo, Barbara},
  title     = {Rethinking Visual Geo-localization for Large-Scale Applications},
  booktitle = {Proc. IEEE/CVF Conf. Comput. Vis. Pattern Recognit. (CVPR)},
  year      = {2022},
  pages     = {4868--4878},
  doi       = {10.1109/CVPR52688.2022.00483}
}

@inproceedings{mixvpr,
  author    = {Ali-bey, Amar and Chaib-draa, Brahim and Gigu\`{e}re, Philippe},
  title     = {{MixVPR}: Feature Mixing for Visual Place Recognition},
  booktitle = {Proc. IEEE/CVF Winter Conf. Appl. Comput. Vis. (WACV)},
  year      = {2023},
  pages     = {2997--3006},
  doi       = {10.1109/WACV56688.2023.00301}
}

@inproceedings{eigenplaces,
  author    = {Berton, Gabriele and Trivigno, Gabriele and Caputo, Barbara and
               Masone, Carlo},
  title     = {{EigenPlaces}: Training Viewpoint Robust Models for Visual Place Recognition},
  booktitle = {Proc. IEEE/CVF Int. Conf. Comput. Vis. (ICCV)},
  year      = {2023},
  pages     = {11046--11056},
  doi       = {10.1109/ICCV51070.2023.01017}
}

@article{anyloc,
  author  = {Keetha, Nikhil and Mishra, Avneesh and Karhade, Jay and
             Jatavallabhula, Krishna Murthy and Scherer, Sebastian and
             Krishna, Madhava and Garg, Sourav},
  title   = {{AnyLoc}: Towards Universal Visual Place Recognition},
  journal = IEEE_J_RAL,
  year    = {2024},
  volume  = {9},
  number  = {2},
  pages   = {1286--1293},
  doi     = {10.1109/LRA.2023.3343602}
}

@inproceedings{salad,
  author    = {Izquierdo, Sergio and Civera, Javier},
  title     = {Optimal Transport Aggregation for Visual Place Recognition},
  booktitle = {Proc. IEEE/CVF Conf. Comput. Vis. Pattern Recognit. (CVPR)},
  year      = {2024},
  pages     = {17658--17668},
  doi       = {10.1109/CVPR52733.2024.01672}
}

@inproceedings{boq,
  author    = {Ali-bey, Amar and Chaib-draa, Brahim and Gigu\`{e}re, Philippe},
  title     = {{BoQ}: A Place is Worth a Bag of Learnable Queries},
  booktitle = {Proc. IEEE/CVF Conf. Comput. Vis. Pattern Recognit. (CVPR)},
  year      = {2024},
  pages     = {17794--17803},
  doi       = {10.1109/CVPR52733.2024.01685}
}

@inproceedings{megaloc,
  author    = {Berton, Gabriele and Masone, Carlo},
  title     = {{MegaLoc}: One Retrieval to Place Them All},
  booktitle = {Proc. IEEE/CVF Conf. Comput. Vis. Pattern Recognit. Workshops (CVPRW)},
  year      = {2025},
  pages     = {2852--2858},
  doi       = {10.1109/CVPRW67362.2025.00269}
}

@inproceedings{fol,
  author    = {Wang, Changwei and Chen, Shunpeng and Song, Yukun and Xu, Rongtao and
               Zhang, Zherui and Zhang, Jiguang and Yang, Haoran and Zhang, Yu and
               Fu, Kexue and Du, Shide and Xu, Zhiwei and Gao, Longxiang and
               Guo, Li and Xu, Shibiao},
  title     = {Focus on Local: Finding Reliable Discriminative Regions for
               Visual Place Recognition},
  booktitle = {Proc. AAAI Conf. Artif. Intell. (AAAI)},
  year      = {2025},
  volume    = {39},
  number    = {7},
  pages     = {7536--7544},
  doi       = {10.1609/aaai.v39i7.32811}
}

@inproceedings{image,
  author    = {Lu, Feng and Jin, Tong and Ye, Canming and Liu, Yunpeng and
               Lan, Xiangyuan and Yuan, Chun},
  title     = {Towards Implicit Aggregation: Robust Image Representation for
               Place Recognition in the Transformer Era},
  booktitle = {Adv. Neural Inf. Process. Syst. (NeurIPS)},
  year      = {2025},
  note      = {arXiv:2511.06024}
}

@inproceedings{sage,
  author    = {Chen, Shunpeng and Wang, Changwei and Xu, Rongtao and Pei, Xingtian and
               Song, Yukun and Lin, Jinzhou and Xu, Wenhao and Zhang, Jingyi and
               Guo, Li and Xu, Shibiao},
  title     = {{SAGE}: Spatial-visual Adaptive Graph Exploration for Efficient
               Visual Place Recognition},
  booktitle = {Proc. Int. Conf. Learn. Represent. (ICLR)},
  year      = {2026},
  note      = {arXiv:2509.25723}
}

@article{dinov2,
  author  = {Oquab, Maxime and Darcet, Timoth\'{e}e and Moutakanni, Th\'{e}o and
             Vo, Huy and Szafraniec, Marc and Khalidov, Vasil and Fernandez, Pierre and
             Haziza, Daniel and Massa, Francisco and El-Nouby, Alaaeldin and
             Assran, Mahmoud and Ballas, Nicolas and Galuba, Wojciech and
             Howes, Russell and Huang, Po-Yao and Li, Shang-Wen and Misra, Ishan and
             Rabbat, Michael and Sharma, Vasu and Synnaeve, Gabriel and Xu, Hu and
             Jegou, Herv\'{e} and Mairal, Julien and Labatut, Patrick and
             Joulin, Armand and Bojanowski, Piotr},
  title   = {{DINOv2}: Learning Robust Visual Features without Supervision},
  journal = {Trans. Mach. Learn. Res. (TMLR)},
  year    = {2024},
  note    = {arXiv:2304.07193}
}

@inproceedings{airloop,
  author    = {Gao, Dasong and Wang, Chen and Scherer, Sebastian},
  title     = {{AirLoop}: Lifelong Loop Closure Detection},
  booktitle = {Proc. IEEE Int. Conf. Robot. Autom. (ICRA)},
  year      = {2022},
  pages     = {10664--10671},
  doi       = {10.1109/ICRA46639.2022.9811658}
}

@article{viper,
  author  = {Ming, Yuhang and Xu, Minyang and Yang, Xingrui and Ye, Weicai and
             Wang, Weihan and Peng, Yong and Dai, Weichen and Kong, Wanzeng},
  title   = {{VIPeR}: Visual Incremental Place Recognition With Adaptive Mining
             and Continual Learning},
  journal = IEEE_J_RAL,
  year    = {2025},
  volume  = {10},
  number  = {3},
  pages   = {3038--3045},
  doi     = {10.1109/LRA.2025.3539093}
}

@inproceedings{incloud,
  author    = {Knights, Joshua and Moghadam, Peyman and Ramezani, Milad and
               Sridharan, Sridha and Fookes, Clinton},
  title     = {{InCloud}: Incremental Learning for Point Cloud Place Recognition},
  booktitle = {Proc. IEEE/RSJ Int. Conf. Intell. Robots Syst. (IROS)},
  year      = {2022},
  pages     = {8559--8566},
  doi       = {10.1109/IROS47612.2022.9981252}
}

@article{ewc,
  author  = {Kirkpatrick, James and Pascanu, Razvan and Rabinowitz, Neil and
             Veness, Joel and Desjardins, Guillaume and Rusu, Andrei A. and
             Milan, Kieran and Quan, John and Ramalho, Tiago and
             Grabska-Barwinska, Agnieszka and Hassabis, Demis and Clopath, Claudia and
             Kumaran, Dharshan and Hadsell, Raia},
  title   = {Overcoming catastrophic forgetting in neural networks},
  journal = {Proc. Nat. Acad. Sci. USA},
  year    = {2017},
  volume  = {114},
  number  = {13},
  pages   = {3521--3526},
  doi     = {10.1073/pnas.1611835114}
}

@inproceedings{si,
  author    = {Zenke, Friedemann and Poole, Ben and Ganguli, Surya},
  title     = {Continual Learning Through Synaptic Intelligence},
  booktitle = {Proc. Int. Conf. Mach. Learn. (ICML)},
  year      = {2017},
  pages     = {3987--3995}
}

@article{lwf,
  author  = {Li, Zhizhong and Hoiem, Derek},
  title   = {Learning without Forgetting},
  journal = IEEE_J_PAMI,
  year    = {2018},
  volume  = {40},
  number  = {12},
  pages   = {2935--2947},
  doi     = {10.1109/TPAMI.2017.2773081}
}

@article{gem_pool,
  author  = {Radenovi\'{c}, Filip and Tolias, Giorgos and Chum, Ond\v{r}ej},
  title   = {Fine-Tuning {CNN} Image Retrieval with No Human Annotation},
  journal = IEEE_J_PAMI,
  year    = {2019},
  volume  = {41},
  number  = {7},
  pages   = {1655--1668},
  doi     = {10.1109/TPAMI.2018.2846566}
}

@inproceedings{msls,
  author    = {Warburg, Frederik and Hauberg, S{\o}ren and Lopez-Antequera, Manuel and
               Gargallo, Pau and Kuang, Yubin and Civera, Javier},
  title     = {Mapillary Street-Level Sequences: A Dataset for Lifelong Place Recognition},
  booktitle = {Proc. IEEE/CVF Conf. Comput. Vis. Pattern Recognit. (CVPR)},
  year      = {2020},
  pages     = {2623--2632},
  doi       = {10.1109/CVPR42600.2020.00270}
}

@inproceedings{inloc,
  author    = {Taira, Hajime and Okutomi, Masatoshi and Sattler, Torsten and
               Cimpoi, Mircea and Pollefeys, Marc and Sivic, Josef and
               Pajdla, Tomas and Torii, Akihiko},
  title     = {{InLoc}: Indoor Visual Localization with Dense Matching and View Synthesis},
  booktitle = {Proc. IEEE/CVF Conf. Comput. Vis. Pattern Recognit. (CVPR)},
  year      = {2018},
  pages     = {7199--7209},
  doi       = {10.1109/CVPR.2018.00752}
}

@inproceedings{baidumall,
  author    = {Sun, Xun and Xie, Yuanfan and Luo, Pei and Wang, Liang},
  title     = {A Dataset for Benchmarking Image-Based Localization},
  booktitle = {Proc. IEEE Conf. Comput. Vis. Pattern Recognit. (CVPR)},
  year      = {2017},
  pages     = {5641--5649},
  doi       = {10.1109/CVPR.2017.598}
}

@inproceedings{scannetpr,
  author    = {Ming, Yuhang and Yang, Xingrui and Zhang, Guofeng and Calway, Andrew},
  title     = {{CGiS-Net}: Aggregating Colour, Geometry and Implicit Semantic
               Features for Indoor Place Recognition},
  booktitle = {Proc. IEEE/RSJ Int. Conf. Intell. Robots Syst. (IROS)},
  year      = {2022},
  pages     = {6991--6997},
  doi       = {10.1109/IROS47612.2022.9981113}
}

@inproceedings{naverindoor,
  author    = {Lee, Donghwan and Ryu, Soohyun and Yeon, Suyong and Lee, Yonghan and
               Kim, Deokhwa and Han, Cheolho and Cabon, Yohann and Weinzaepfel, Philippe and
               Guerin, Nicolas and Csurka, Gabriela and Humenberger, Martin},
  title     = {Large-scale Localization Datasets in Crowded Indoor Spaces},
  booktitle = {Proc. IEEE/CVF Conf. Comput. Vis. Pattern Recognit. (CVPR)},
  year      = {2021},
  pages     = {3226--3235},
  doi       = {10.1109/CVPR46437.2021.00324}
}

@inproceedings{nycindoorvpr,
  author    = {Sheng, Diwei and Yang, Anbang and Rizzo, John-Ross and Feng, Chen},
  title     = {{NYC-Indoor-VPR}: A Long-Term Indoor Visual Place Recognition Dataset
               with Semi-Automatic Annotation},
  booktitle = {Proc. IEEE Int. Conf. Robot. Autom. (ICRA)},
  year      = {2024},
  note      = {arXiv:2404.00504}
}

@inproceedings{textinplace,
  author    = {Tao, Huaqi and Liu, Bingxi and Chen, Calvin and Huang, Tingjun and
               Li, He and Cui, Jinqiang and Zhang, Hong},
  title     = {{TextInPlace}: Indoor Visual Place Recognition in Repetitive Structures
               with Scene Text Spotting and Verification},
  booktitle = {Proc. IEEE/RSJ Int. Conf. Intell. Robots Syst. (IROS)},
  year      = {2025},
  note      = {arXiv:2503.06501}
}

@article{gsvcities,
  author  = {Ali-bey, Amar and Chaib-draa, Brahim and Gigu\`{e}re, Philippe},
  title   = {{GSV-Cities}: Toward appropriate supervised visual place recognition},
  journal = {Neurocomputing},
  year    = {2022},
  volume  = {513},
  pages   = {194--203},
  doi     = {10.1016/j.neucom.2022.09.127}
}

@inproceedings{tartanair,
  author    = {Wang, Wenshan and Zhu, Delong and Wang, Xiangwei and Hu, Yaoyu and
               Qiu, Yuheng and Wang, Chen and Hu, Yafei and Kapoor, Ashish and
               Scherer, Sebastian},
  title     = {{TartanAir}: A Dataset to Push the Limits of Visual {SLAM}},
  booktitle = {Proc. IEEE/RSJ Int. Conf. Intell. Robots Syst. (IROS)},
  year      = {2020},
  pages     = {4909--4916},
  doi       = {10.1109/IROS45743.2020.9341801}
}

@article{robotcar,
  author  = {Maddern, Will and Pascoe, Geoffrey and Linegar, Chris and Newman, Paul},
  title   = {1 year, 1000 km: The {Oxford RobotCar} dataset},
  journal = {Int. J. Robot. Res.},
  year    = {2017},
  volume  = {36},
  number  = {1},
  pages   = {3--15},
  doi     = {10.1177/0278364916679498}
}

@article{lowry_survey,
  author  = {Lowry, Stephanie and S\"{u}nderhauf, Niko and Newman, Paul and
             Leonard, John J. and Cox, David and Corke, Peter and Milford, Michael J.},
  title   = {Visual Place Recognition: A Survey},
  journal = IEEE_J_RO,
  year    = {2016},
  volume  = {32},
  number  = {1},
  pages   = {1--19},
  doi     = {10.1109/TRO.2015.2496823}
}

@article{vpr_tutorial,
  author  = {Schubert, Stefan and Neubert, Peer and Garg, Sourav and Milford, Michael and
             Fischer, Tobias},
  title   = {Visual Place Recognition: A Tutorial},
  journal = IEEE_M_RA,
  year    = {2024},
  volume  = {31},
  number  = {3},
  pages   = {139--153},
  doi     = {10.1109/MRA.2023.3310859}
}

@inproceedings{patchnetvlad,
  author    = {Hausler, Stephen and Garg, Sourav and Xu, Ming and Milford, Michael and
               Fischer, Tobias},
  title     = {{Patch-NetVLAD}: Multi-Scale Fusion of Locally-Global Descriptors
               for Place Recognition},
  booktitle = {Proc. IEEE/CVF Conf. Comput. Vis. Pattern Recognit. (CVPR)},
  year      = {2021},
  pages     = {14136--14147},
  doi       = {10.1109/CVPR46437.2021.01392}
}

@inproceedings{effovpr,
  author    = {Tzachor, Issar and Lerner, Boaz and Levy, Matan and Green, Michael and
               Shalev, Tal Berkovitz and Habib, Gavriel and Samuel, Dvir and
               Zailer, Noam Korngut and Shimshi, Or and Darshan, Nir and Ben-Ari, Rami},
  title     = {{EffoVPR}: Effective Foundation Model Utilization for
               Visual Place Recognition},
  booktitle = {Proc. Int. Conf. Learn. Represent. (ICLR)},
  year      = {2025},
  note      = {arXiv:2405.18065}
}

@article{nordland,
  author  = {Olid, Daniel and F\'{a}cil, Jos\'{e} M. and Civera, Javier},
  title   = {Single-View Place Recognition under Seasonal Changes},
  journal = {arXiv:1808.06516},
  year    = {2018},
  note    = {PPNIV Workshop, IROS 2018}
}

@inproceedings{apriltag,
  author    = {Olson, Edwin},
  title     = {{AprilTag}: A robust and flexible visual fiducial system},
  booktitle = {Proc. IEEE Int. Conf. Robot. Autom. (ICRA)},
  year      = {2011},
  pages     = {3400--3407},
  doi       = {10.1109/ICRA.2011.5979561}
}

@inproceedings{gem,
  author    = {Lopez-Paz, David and Ranzato, Marc'Aurelio},
  title     = {Gradient Episodic Memory for Continual Learning},
  booktitle = {Adv. Neural Inf. Process. Syst. (NeurIPS)},
  year      = {2017},
  note      = {arXiv:1706.08840}
}

@inproceedings{er,
  author    = {Rolnick, David and Ahuja, Arun and Schwarz, Jonathan and
               Lillicrap, Timothy P. and Wayne, Greg},
  title     = {Experience Replay for Continual Learning},
  booktitle = {Adv. Neural Inf. Process. Syst. (NeurIPS)},
  year      = {2019},
  note      = {arXiv:1811.11682}
}

@article{umeyama,
  author  = {Umeyama, Shinji},
  title   = {Least-squares estimation of transformation parameters between
             two point patterns},
  journal = IEEE_J_PAMI,
  year    = {1991},
  volume  = {13},
  number  = {4},
  pages   = {376--380},
  doi     = {10.1109/34.88573}
}

@inproceedings{ate_tum,
  author    = {Sturm, J{\"u}rgen and Engelhard, Nikolas and Endres, Felix and
               Burgard, Wolfram and Cremers, Daniel},
  title     = {A benchmark for the evaluation of {RGB-D} {SLAM} systems},
  booktitle = {Proc. IEEE/RSJ Int. Conf. Intell. Robots Syst. (IROS)},
  year      = {2012},
  pages     = {573--580},
  doi       = {10.1109/IROS.2012.6385773}
}

@article{yu2023brain,
  title={Brain-inspired multimodal hybrid neural network for robot place recognition},
  author={Yu, Fangwen and Wu, Yujie and Ma, Songchen and Xu, Mingkun and Li, Hongyi and Qu, Huanyu and Song, Chenhang and Wang, Taoyi and Zhao, Rong and Shi, Luping},
  journal={Science Robotics},
  volume={8},
  number={78},
  pages={eabm6996},
  year={2023},
  publisher={American Association for the Advancement of Science},
  doi = {10.1126/scirobotics.abm6996}
}

\end{document}